\documentclass[runningheads]{llncs}
\usepackage[T1]{fontenc}
\usepackage{graphicx}
\usepackage{amsmath,amssymb} % define this before the line numbering.
\usepackage{subcaption}
\usepackage{wrapfig}
\usepackage{hyperref}
\usepackage[capitalize,noabbrev]{cleveref}
\usepackage{algorithm, algorithmic}
\usepackage{booktabs} % for professional tables
\graphicspath{{figures/}}
\usepackage{color}

\begin{document}
\title{Neural Network Panning: Screening the Optimal Sparse Network Before Training}
\titlerunning{Neural Network Panning}
\author{Xiatao Kang\inst{1} \and
Ping Li\inst{1,3,*} \and
Jiayi Yao\inst{1} \and
Chengxi Li\inst{2}}
\authorrunning{X. Kang et al.}
\institute{School of Computer and Communication Engineering, Changsha University of Science and Technology \and
School of Computer, Xidian University \and
Hunan Provincial Key Laboratory of Intelligent Processing of Big Data on Transp \\
{\tt\small kangxiatao@gmail.com},
{\tt\small lping9188@163.com}
}
\maketitle              % typeset the header of the contribution
\begin{abstract}
Pruning on neural networks before training not only compresses the original models, but also accelerates the network training phase, which has substantial application value. The current work focuses on fine-grained pruning, which uses metrics to calculate weight scores for weight screening, and extends from the initial single-order pruning to iterative pruning. Through these works, we argue that network pruning can be summarized as an expressive force transfer process of weights, where the reserved weights will take on the expressive force from the removed ones for the purpose of maintaining the performance of original networks. In order to achieve optimal expressive force scheduling, we propose a pruning scheme before training called Neural Network Panning which guides expressive force transfer through multi-index and multi-process steps, and designs a kind of panning agent based on reinforcement learning to automate processes. Experimental results show that Panning performs better than various available pruning before training methods. Our code is made public at: \href{https://github.com/kangxiatao/RLPanning}{https://github.com/kangxiatao/RLPanning.}

\end{abstract}
%
%
%
%===========================================================
\section{Introduction}

In recent years, neural networks have achieved breakthrough results in computer vision\cite{Alexnet2012,ObjectDetection2019} and natural language processing\cite{BERT2020}. More complex architectural designs improve model performance but significantly increase the number of parameters of the neural network. Research on neural network compression\cite{Compression2016} is devoted to reducing the memory overhead and computational cost of neural networks and maintaining the performance of the network architecture. Neural network pruning\cite{Prune2015} is the most direct method of neural network compression, and the purpose of compression is achieved by filtering out unimportant weights.

The latest research focuses on network pruning before training\cite{Lee20Signal,Robust2021,Missing}, with progress in theory and application. A series of studies of the lottery ticket hypothesis\cite{Lottery,frankle2019stabilizing,Malach2020Proving} demonstrates the effectiveness of pruning before training. Many methods of pruning before training achieve desirable performance\cite{Jorge21Trimming,SynFlow,Connectivity21}. However, some work argues that pruning before training should be considered a search of the network architecture\cite{Rethinking19} and that random selection of weights with equal layer pruning quotas\cite{Missing,RandomTickets} can even achieve better generalization. Drop weights randomly or by structure from left to right during training using sparse constraints\cite{DAM21}, and both are very effective. We believe that the idea of these works is that weights are equal, and the network structure is directly examined when pruning the network rather than the weight ordering under a particular index.

\begin{figure}[t]
    \begin{center}
    \centerline{\includegraphics[width=\columnwidth]{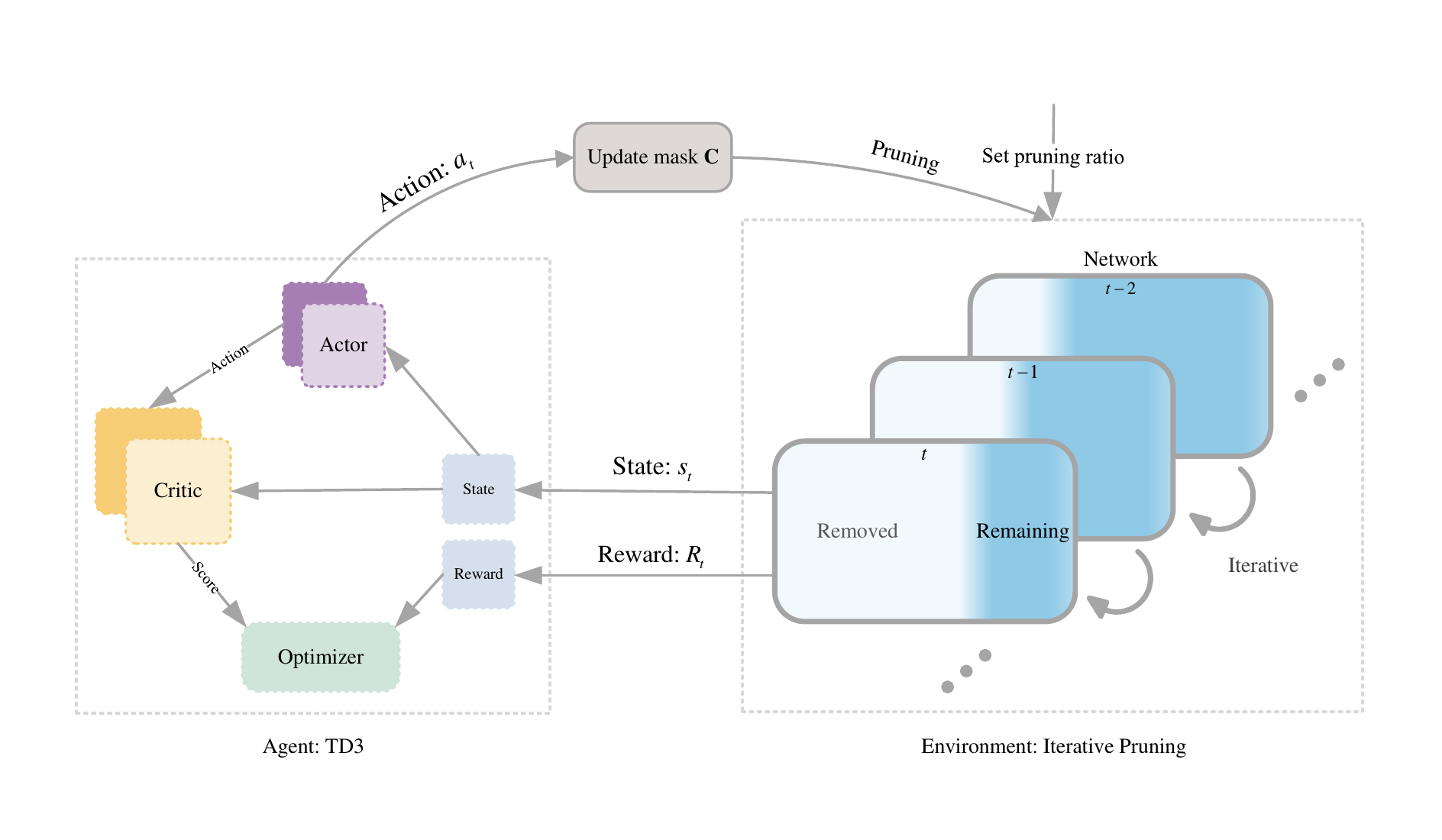}}
    \caption{Overview of Panning based on reinforcement learning. On the right is the Panning environment, which iteratively prunes the network after setting the pruning rate, and the feedback includes the current state of the sparse network and the set reward. On the left is the TD3 agent, which acts through the feedback given by the environment, that is, it selects the optimal pruning strategy based on the network state during iterative pruning.}
    \label{rl_pruning}
    \end{center}
  \end{figure}

The characterization of weight expressiveness has always been the research focus of neural network pruning algorithms. Using sensitivity\cite{SNIP} as a metric is currently the mainstream method for pruning before training. In our research, we found that the process of pruning can be regarded as the process of transferring the expressive force of the removed weight to the remaining weight so that any weight has an equal opportunity to express. Assuming that the weight sensitivity is a representation of the weight expressive force, \cref{force_transfer} shows the expressive force transfer through iterative pruning. Different measurement methods indicate different affordability of weights, and a single measurement index is always flawed. Loss-sensitive correlation metrics\cite{SNIP,GraSP} are prone to inter-layer disconnection at high compression rates, while metrics that only consider inter-layer equilibrium\cite{SynFlow} do not generalize well at low compression rates. Some methods improve performance using iterative pruning with a single metric and adding constraints\cite{ESPN,Connectivity21}, but this benefits from the fact that iterative pruning examines structural changes and does not improve the one-sidedness of the single metric.

Therefore, we propose a multi-metric, multi-process progressive pruning method based on our improvements and existing metrics. The process of picking weights is like a gold purification process, which we call Panning. Since dynamic iterative pruning will generate many state change trajectories, we design Panning as an environment with actions and feedback, and use reinforcement learning to search for Panning steps(called RLPanning), that is, the agent automates the Panning process through policy learning(\cref{rl_pruning}). Unlike simple metric ranking, RLPanning examines multiple expressive forces during pruning and determines whether and when the forces can be transferred according to the state of the target network in terms of loss variation, sparsity, and interlayer structure to maximize the benefits and obtain the ideal sparse network.

The contributions of this paper are as follows:
\begin{itemize}
    \item We generalize the pruning process as the expressive force transfer process, analyze various weight metrics, and make improvements.
    \item We propose a multi-metric, multi-process iterative pruning method before training, called Panning. Panning performs well by pruning the network before training, maintaining generalization well under extremely high compression rates.
    \item We design a Panning environment and use reinforcement learning to learn a Panning strategy by sampling spatial actions, called RLPanning. RLPanning obtains the optimal pruning strategy with a comprehensive measure of multiple indicators through dynamic expressive force transfer.
\end{itemize}

%===========================================================
\section{Related work}

Neural network pruning has been a hot research topic along with the development of artificial intelligence so far. The purpose of pruning is to reduce the parameters and computation of neural networks to achieve compression of neural networks\cite{Compression2016}, which can be divided into two major categories of unstructured and structured pruning according to the way of pruning networks. Unstructured pruning performs fine-grained pruning intending to minimize the number of parameters, and the classical process is training, pruning, and fine-tuning\cite{Prune2015}. Sparse constraints in training to obtain sparse structures are also more widely used\cite{ToPrune18,Scalable2018}. Structured pruning focuses on the structure of the network, and the pruning results in a lean, compact model. In convolutional networks, structured pruning is usually considered the pruning of the filter\cite{PruningFilters17}, and most of the work sets structured constraints\cite{VACL19,LearningEfficient17,LearningStructured16} in training to prune the network.

In order to obtain better compression structure and performance, many works are devoted to automatic neural network architecture search\cite{LargeScale17,Learning18,ArchitectureSearch17}. Although automatic search network architecture is prevalent, it also suffers from substantial computational overhead due to the infinite search space. Subsequent work has made new attempts to use adversarial learning\cite{Towards19} or reinforcement learning\cite{AMC18} to search for pruning strategies under a fixed framework to achieve network simplification.

\paragraph{Pruning before training.} 

Pruning before training has substantial theoretical and applied value. The lottery ticket hypothesis\cite{Lottery} demonstrates the possibility of obtaining superior sparse networks before training through extensive experiments. Since the lottery hypothesis was put forward, much work has been done on expansion\cite{desai2019evaluating,frankle2019stabilizing,morcos2019oneticket} and theoretical research\cite{Malach2020Proving,Logarithmic20}.

In the current application, the pruning before training focuses on fine-grain pruning, and the weight is usually measured by examining the effect of removing a certain weight on a certain state of the network. Specifically, SNIP\cite{SNIP} examines the impact of pruning weights on loss, that is, the sensitivity of weights to model losses. GraSP\cite{GraSP} examines the degree to which the removal weight decreases the loss, that is, the sensitivity of the weight to the change in loss. SynFlow\cite{SynFlow} starts from the trainability of the maximum compression network and examines the balance of weights between layers. FORCE\cite{Jorge21Trimming} progressively prunes the network based on the SNIP method and allows for the resurrection of removed weights, taking into account changes in the network structure. Among them, SNIP and GraSP methods can obtain sparse networks with only a single network pruning. Methods that consider inter-layer balance and network structure require iterative network pruning. In addition, there are iterative pruning methods to make dynamic expansion\cite{ESPN,OverParame21} or set constraints\cite{Connectivity21} to obtain good performance.

For weight measurement, SynFlow summarizes the mathematical expression of sensitivity, called synaptic saliency\cite{SynFlow}$\mathcal{S}(\theta)=\frac{\partial\mathcal{R}}{\partial\theta}\odot\theta$, where $\mathcal{R}$ is a scalar loss function, and $\theta$ is a parameter to be examined. Hence $\mathcal{S}_{SNIP}(\theta)=\left|\frac{\partial\mathcal{L}}{\partial\theta}\odot\theta\right|,\mathcal{S}_{GraSP}(\theta)=\frac{\partial g^Tg}{\partial\theta}\odot\theta$, $g^Tg$ is the first-order Taylor approximation of loss drop.

%===========================================================
\section{Methodology}

In this section, we introduce Panning in detail and then automate Panning using reinforcement learning. Our goal is to have the agent schedule the expressive force of the weights according to the state of the network, cross-validate the weights with a variety of meaningful metrics, and find the optimal sparse sub-network before training. Finally, we discuss and improve the existing metrics and apply them in Panning.

%-----------------------------------------------------------
\subsection{Problem definition}

We mainly consider fine-grained pruning of neural networks, that is, removing parts of weights that are considered unimportant, regardless of the structure of the network. In order to describe this irregular sparse network conveniently, it is usually represented by the Hadamard product of the mask $\mathbf{c}$ and the weight $\mathbf{w}$, then the optimization problem of the network is:
\begin{equation}
  \begin{gathered}
  \min _{\mathbf{c}, \mathbf{w}} \mathcal{L}(\mathbf{c} \odot \mathbf{w} ; \mathcal{D}) \\
  \text { s.t. } \mathbf{c} \in\{0,1\}^{m},\|\mathbf{c}\|_{0} \leq m(1-\rho)
  \end{gathered}
\end{equation}
Where $\mathcal{L}(\cdot)$ represents the loss function, $\mathcal{D}$ is the dataset, $m$ is the total number of weights in the network, and $\rho$ is the target pruning ratio (sparse degree). When the mask $\mathbf{c}$ corresponding to the weight is set to zero, it means that the weight is removed.

%-----------------------------------------------------------
\subsection{Transfer of expressive force}

\begin{figure}[h]
    \centering
    \begin{subfigure}{0.48\columnwidth}
    % The figure in PDF is too large, which affects the loading of the article. Use compressed PNG.
    \includegraphics[width=\columnwidth]{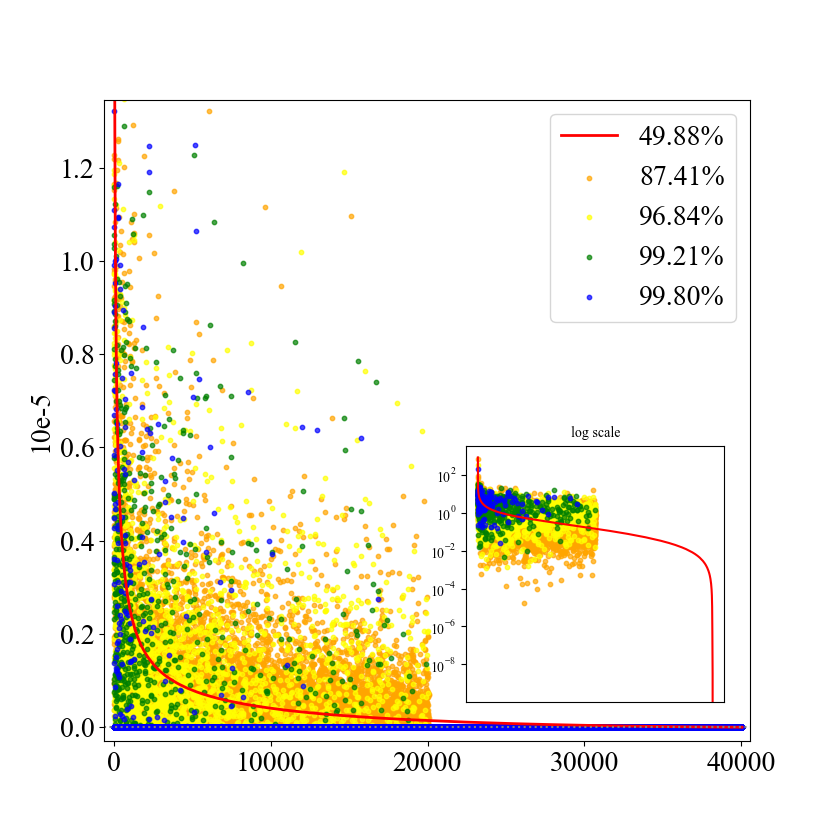}  
    \end{subfigure}
    \begin{subfigure}{0.48\columnwidth}
    \includegraphics[width=\columnwidth]{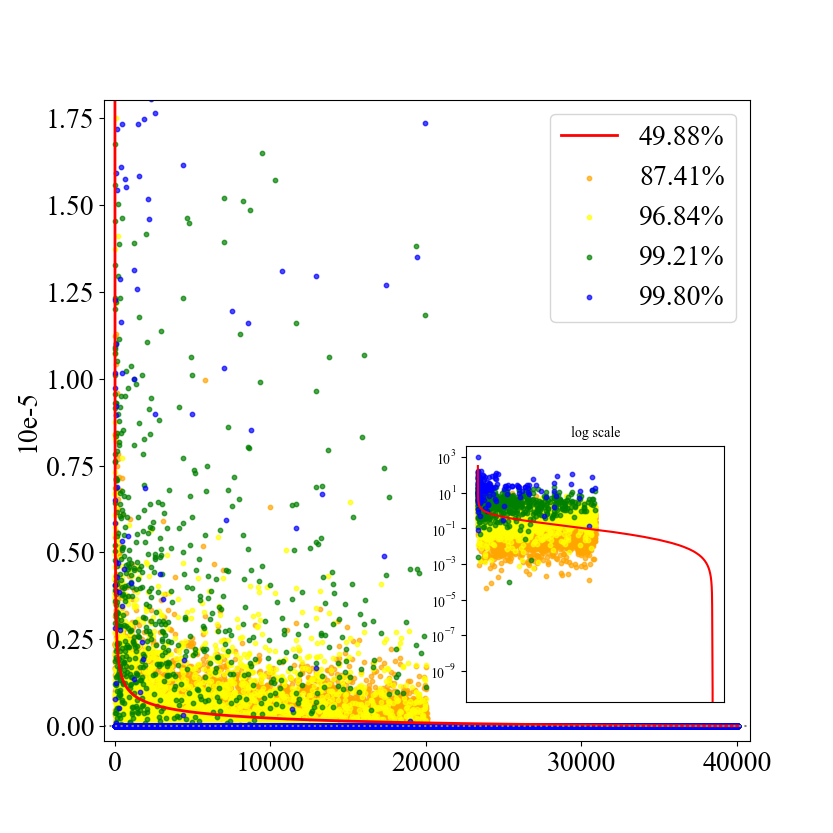}
    \end{subfigure}
    \caption{Comparison of metric scores at different pruning ratios during the iterative pruning process of SNIP (left) and GraSP (right). Experiments were performed under VGG19/CIFAR10. The vertical axis is the metric score $\mathcal{S}$ (where $\mathcal{S}_{GraSP}$ is the absolute value), and the horizontal axis is the number of weights (in order to reduce the amount of data, after flattening all weights into vectors, take a value every 500 weights). After sorting the weight scores for the first time (line), the score changes (dots) under other pruning ratios are displayed at the corresponding weight positions. The subplots are the logarithmic changes in the vertical axis scale.}
    \label{force_transfer}
\end{figure}

We use SNIP and GraSP for iterative pruning, respectively, and show in \cref{force_transfer} the changes in the metric scores of all weights in the network under different compression ratios (the two adjacent pruning rates can represent the change in scores before and after pruning), and more obvious hierarchical changes can be observed after taking the logarithm of the vertical axis (as shown in the subplot). We find that the weight expressive force is sensitive to the compression ratio. As the compression ratio increases, some of the less expressive weight metrics improve significantly (above the red line), and vice versa. 

Iterative dynamic pruning is widely considered to perform better, especially before training. Through continuous compression and gradient iterations, weights that were evaluated as strong in performance in the most recent iteration are retained. The invariant properties of flow transported between layers are given in the SynFLow\cite{SynFlow}. Naturally, we view model compression as a process of transferring expressive force between different weights based on flow sensitivity. Take $\mathcal{S}(\theta)=|\frac{\partial\mathcal{R}}{\partial\theta}\odot\theta|$ as the weight expressive force, and only observe the magnitude of the force.

Specifically, the expressive force of the removed weight will be transferred to the remaining weight, and the force of the remaining part will also be adjusted due to structural changes. Therefore, weights with low importance at the beginning may become critical weights in the future, and a single pruning can easily remove this part of the weights, resulting in the unreasonable distribution of weight expressive force. It is challenging to maintain layer balance even if layer collapse does not occur. The advantage of iterative pruning is that it gradually transfers the expressive force and examines the changes in the structure so that the connection structure between the remaining weights is better maintained.

%-----------------------------------------------------------
\subsection{Artificial Panning}

Although iterative pruning solves the layer collapse problem, existing iterations use a single metric pruning throughout, ignoring the expressive force of weights under other metrics. Simply summing up multiple metrics for measurement will easily cover up specific properties, resulting in an indistinct degree of weight discrimination. We design the Panning algorithm based on dynamic iterations to establish multiple mappings between sparse network structure changes and weights. The characteristic of Panning is to observe the state of the network in the iterative process, and select the most favorable indicator to measure the weight according to the state change. Based on this, the network is progressively pruned in a loop, and the removed weights are allowed to re-live until the desired sparse network is finally obtained.

Specifically, the pruning rate changes exponentially during the iteration process, and the pruning ratio $\rho_i$ of round $i$ is:
\begin{equation}
  \rho_i=1-\left(1-\rho_{target}\right)^{\left(\frac{i}{T}\right)}
\end{equation}
where $\rho_{target}$ is the target pruning ratio, and $T$ is the total the number of iterations. Then calculate the various metrics in the current state of the model, and obtain the score of the weight $\mathbf{w}$ through a specific fusion:
\begin{equation}
  \mathcal{S}_\mathbf{w}=\sum_{i=1}^{k}{p_i\mathcal{N}\left(\mathcal{S}_{i}\right)}
\end{equation}
where $p_i$ is a hyperparameter that adjusts the proportion of the metric, and $\mathcal{N}(\mathcal{S}_{i})$ represents the normalization of the metric score. Finally, the network is gradually pruned according to the score $\mathcal{S}_\mathbf{w}$ and the pruning ratio $\rho_i$: 
\begin{equation}
  \mathbf{c}=\ \mathcal{S}_\mathbf{w}(\boldsymbol{\theta})>\mathcal{S}_{top(m(1-\rho_i))}
\end{equation}
where $\mathcal{S}_{top(\kappa)}$ is the $\kappa$-th value after the descending order, and $m$ is the total number of weights. When the weight $\mathbf{w}_i$ does not satisfy the conditions of the above formula, the corresponding mask $\mathbf{c}_{i}$ is set to zero, the weight is removed and exists as a zero value. The value of $\mathbf{w}_i$ is reset the next time the weight sensitivity $\mathcal{S}$ is calculated, so that the removed weight can be restored. \cref{alorithm_panning} shows the complete gold panning steps.

\begin{algorithm}[tb]
  \caption{Panning}
  \label{alorithm_panning}
  \begin{algorithmic}[1]
      \REQUIRE Weight $\mathbf{w}$, batch data $\mathcal{D}^{b}$, target pruning rate $\rho_{target}$.
      % \RETURN $\mathbf{c}$

      \STATE Initialize $\mathbf{w}$, $\mathbf{c}$
      \FOR{$i=0$ {\bfseries to} $T$}
        \STATE Remove weight $\mathbf{w}^{\ast}=\mathbf{c} \odot \mathbf{w}$ 
        \STATE Calculate loss $\mathcal{R}$
        \STATE Reset weight $\mathbf{w}^{\ast}=\mathbf{w}$
        % \STATE Calculate weight sensitivity $\mathcal{S}$
        \STATE Calculate score $\mathcal{S}_\mathbf{w}=\sum_{i=1}^{k}{p_i\mathcal{N}(\mathcal{S}_{i})}$
        \STATE Update pruning rate $\rho_i=1-\left(1-\rho_{target}\right)^{\left(\frac{i}{T}\right)}$
        \STATE Update mask $\mathbf{c}$
      \ENDFOR
      
      \STATE return $\mathbf{c}$
  \end{algorithmic}
\end{algorithm}

%-----------------------------------------------------------
\subsection{Panning based on reinforcement learning}

According to our observation, an ideal sparse network can be obtained by setting the hyperparameters of Panning based only on the compression of the network. Naturally, if more network states are examined and dynamic expressive force transfer is achieved based on state changes, this can theoretically lead to a sparse network with better performance. Therefore, we design Panning as an environment containing action and state space, so that the intelligent body can automate Panning through policy learning. \cref{rl_pruning} gives an overview of RLPanning, which is described in detail in this section.

\paragraph{Panning environment.} 
Each Panning iteration results in a new sparse network, that is, different sparsity and performance. The number of iterations is taken as the time $t$, so the state space $s_t$ of the Panning environment is:
\begin{equation}
  s_{t}=\left(\mathcal{L}, \Delta \mathcal{L}, \mathcal{L}_{s}, \Delta \mathcal{L}_{s}, \rho_{t}, \rho_{e}, t\right)
\end{equation}
where $\mathcal{L}_s,\Delta \mathcal{L}_{s}$ are the loss and loss reduction of the sparse sub-network in the current state, respectively, $\rho_{e}$ represents the effective compression rate of the network\cite {Connectivity21} (Invalid retention due to disconnection between layers is prone to occur at high compression rates). Note that all states are normalized to facilitate agent training. The action space a of Panning is the hyperparameter $p_i$ that regulates the proportion of metrics, and assuming that there are $\kappa$ metrics, the action space dimension is $\kappa$ and is continuous for the action: 
\begin{equation}
  a \in[-1,1], p_i=\frac{a_i+1}{2}
\end{equation}

Our target is to affect $\mathcal{L},\Delta \mathcal{L}$ as little as possible during pruning and to ensure that the effective compression rate is close to the target compression rate. So we design a reward term $R_t$ consisting of four components: 
\begin{equation}\label{reward_eq}
  \begin{aligned}
  &R_{t}=-\left|\mathcal{N}(\mathcal{L})-\mathcal{N}\left(\mathcal{L}_{s}\right)\right|-\left|\mathcal{N}(\Delta \mathcal{L})-\mathcal{N}\left(\Delta \mathcal{L}_{s}\right)\right|-\alpha\left|\rho_{e}-\rho_{t}\right|-r_{\text {done }} \\
  &r_{\text {done }}=\left\{\begin{array}{cl}
  T-t, & \text { if } \rho_{e}=1 \\
  0, & \text { otherwise }
  \end{array}\right.
  \end{aligned}
\end{equation}
where $\mathcal{N}(\cdot)$ represents normalization. $T$ is the maximum number of iterations. $r_{done}$ is the reward in the final state, $done=True$ when $t=T$ or effective compression ratio $\rho_e=1$, $\rho_e=1$ means that all weights are removed, and the iteration will end early. $\alpha$ is the reward strength that adjusts the quality of network compression. Usually the ineffective compression weight is less than one percentage point, and we set $\alpha$ to $100$.

\paragraph{Panning agent.} 
The continuous action of Panning is a deterministic strategy, and we adopt TD3\cite{TD3}, which performs well in the field of continuous control, as the Panning agent. TD3 is an actor-critic network architecture. As an optimized version of DDPG\cite{DDPG}, the delay and smoothing processing of TD3 largely mitigates the effects of sudden changes in environmental states due to different batches of samples, which is well suited to our Panning environment.

The workflow of the TD3 agent is shown on the left in \cref{rl_pruning}. The actor network gives an action, and the critic network estimates the reward for the action. The critic network is equivalent to the Q network in DQN\cite{DQN}, and the goal is to solve the action a that maximizes the Q value. The actions and states of the Panning environment are all standardized, the structure and optimization of the TD3 agent basically do not need to be changed, and the network scale and all hyperparameters are shown in the experimental part.

In addition, after resetting the environment, in order to better sample in the space, the final target compression ratio $\rho_T$ is randomly selected between \{80\%,99.99\%\}. In addition, we have set a curriculum learning\cite{CurriculumLearning18} from easy to complex, that is, the probability of the compression target $\rho_T$ being a low compression ratio in the early stage of the agent training is high, and the probability of $\rho_T > 99\%$ in the later stage is higher, thereby speeding up the convergence speed of the agent training.

%-----------------------------------------------------------
\subsection{Selection of metrics}
\label{metrics}

The currently widely accepted singe-shot pruning metrics are SNIP and GraSP, which are very sample-dependent and perform poorly at high compression ratios. SynFlow is more prominent in iterative pruning and achieves a good balance between layers. The most significant advantage of SynFlow is that it does not require samples, but the lack of consideration of sample characteristics also makes it generalize poorly at low compression rates. Actually, SNIP, GraSP, and SynFlow are complementary and computationally very similar. Therefore, we choose these three metrics to apply to Panning, with action space $\kappa=3$ and weight score $\mathcal{S}_w={p_1\mathcal{N}\left(\mathcal{S}_{SynFlow}\right)+p}_2\mathcal{N}\left(\mathcal{S}_{SNIP}\right)+p_3\mathcal{N}\left(\mathcal{S}_{GraSP}\right)$.

Due to gradient decay in the deep network, when the SNIP pruning ratio is large, it is easy to remove too many subsequent convolutional layers and cause faults. The idea of GraSP is to maintain the gradient flow and prune the weights to maximize the loss drop, which improves compressibility. However, there is a significant error in the calculation of the first-order approximation $g^Tg$, especially in the ill-conditioned problem\cite{goodfellow2016deep} that the first-order term of the loss function is smaller than the second-order term during training. Based on this, we believe that maximizing the gradient norm is not equivalent to maintaining the gradient flow, and it is more accurate to remove weights that have little influence on the gradient flow. Therefore, we changed $\mathcal{S}_{GraSP}$ to $|\frac{\partial g^Tg}{\partial\theta}\odot\theta|$, that is, to examine the impact of weights on model trainability. If it is analyzed from the perspective of acting force, after taking the absolute value, it means that only the magnitude of the force is considered, and the direction of action of the force is ignored. Moreover, the experiment proves that the modified $\mathcal{S}_{GraSP}$ has better performance.

In addition, in the classification task, when $\mathcal{R}$ is the fitting loss $\mathcal{L}$ of the network, the gradient information brought by different samples is more abundant. Therefore, when calculating the loss $\mathcal{L}$, we set a specific sampling to ensure that $\mathcal{D}^b$ contains all categories and an equal number of samples for each category. Suppose the dataset has $l$ categories, $k$ samples are taken from each category, then $\mathcal{D}^b=\left\{\left(x^b,y^b\right)\right\},\ x^b=\left(x_{1,2,\ldots,k}^1,x_{1,2,\ldots,k}^2,\ldots,x_{1,2,\ldots,k}^l\right)$.

%===========================================================
\section{Experiments}
%-----------------------------------------------------------
\subsection{Experimental setup}

In the image classification task, we prune LeNet5, VGG19, and ResNet18 convolutional networks with panning before training and select MNISIT, FashionMNISIT, CIFAR10/100,  and TinyImageNet datasets to evaluate our method. In this experiment, images are enhanced by random flipping and cropping. Details and hyperparameters of sparse network training after pruning are shown in \cref{hyperparameters}.

\begin{table}[h]
  \centering
  \caption{Model optimizer and hyperparameters.}
  \label{hyperparameters}
  \resizebox{0.8\columnwidth}{!}{
  \renewcommand\arraystretch{0.9}
  \begin{tabular}{l cccc}
  \cmidrule[\heavyrulewidth](lr){1-5}
  & \multicolumn{1}{c}{MNIST} & \multicolumn{1}{c}{CIFAR} & \multicolumn{2}{c}{\makebox[0.2\textwidth]{ImageNet}} \\ 
  \cmidrule(lr){2-2} \cmidrule(lr){3-3} \cmidrule(lr){4-5}
  & LeNet & VGG/ResNet & VGG & ResNet \\ 
  \cmidrule(lr){1-5}
  Optimizer & \multicolumn{4}{c}{Momentum (0.9)} \\ 
  Learning Rate & \multicolumn{4}{c}{Cosine Annealing (0.1)} \\ 
  Training Epochs & 80 & 180 & 200 & 300 \\ 
  Batch Size & 256 & 128 & 128 & 128 \\ 
  Weight Decay & 1e-4 & 5e-4 & 5e-4 & 1e-4 \\ 
  \cmidrule(lr){1-5}
  \end{tabular}}
\end{table}

%-----------------------------------------------------------
\subsection{Performance of the improved metric}

The performance comparison between the improved SNIP and GraSP metrics and the original method on VGG19/CIFAR10 is shown in \cref{improved_perf}. The results were obtained by taking the average of three experiments.

Note that the loss is computed using a batch of samples containing all labels, although the overall improved SNIP has only a slight improvement in accuracy, which is essential on datasets with more complex labels. Under the 99\% pruning rate, the test set accuracy is still 10\%, which is caused by the fault of SNIP itself. For the improvement of GraSP, the improvement of accuracy is more prominent, especially after the compression rate is higher than 90\%. This suggests that sensitivity to the gradient norm is better than maximizing the gradient norm, validating our previous analysis.

\begin{table}[t]
  \centering
  \caption{Performance comparison of improved SNIP and GraSP.}
  \label{improved_perf}
  \resizebox{0.8\columnwidth}{!}{
  \begin{tabular}{l ccccc}
  \cmidrule[\heavyrulewidth](lr){1-6}
  \makebox[0.35\columnwidth][l]{\textbf{VGG19/CIFAR10}} & \multicolumn{5}{c}{Acc: 94.20\%} \\ 
  \cmidrule(lr){1-6}
  \textbf{Pruning ratio} & 85\% & 90\% & 95\% & 98\% & 99\% \\ 
  \cmidrule(lr){1-6}
  SNIP & 93.91 &	93.82 &	93.72 &	91.16 &	10.00 \\
  Ours SNIP	& \textbf{94.05} & \textbf{93.98} & \textbf{93.86} & \textbf{91.83} & 10.00 \\
  \cmidrule(lr){1-6}
  GraSP & 93.59 &	93.50 &	92.90 &	92.39 &	91.04 \\
  Ours GraSP	& \textbf{93.77} & \textbf{93.69} & \textbf{93.48} & \textbf{92.78} & \textbf{91.84} \\
  \cmidrule(lr){1-6}
  \end{tabular}}
\end{table}

\begin{figure*}[t]
  \centering
  \minipage{0.49\textwidth}
    \begin{subfigure}{\textwidth}
    \includegraphics[width=\columnwidth]{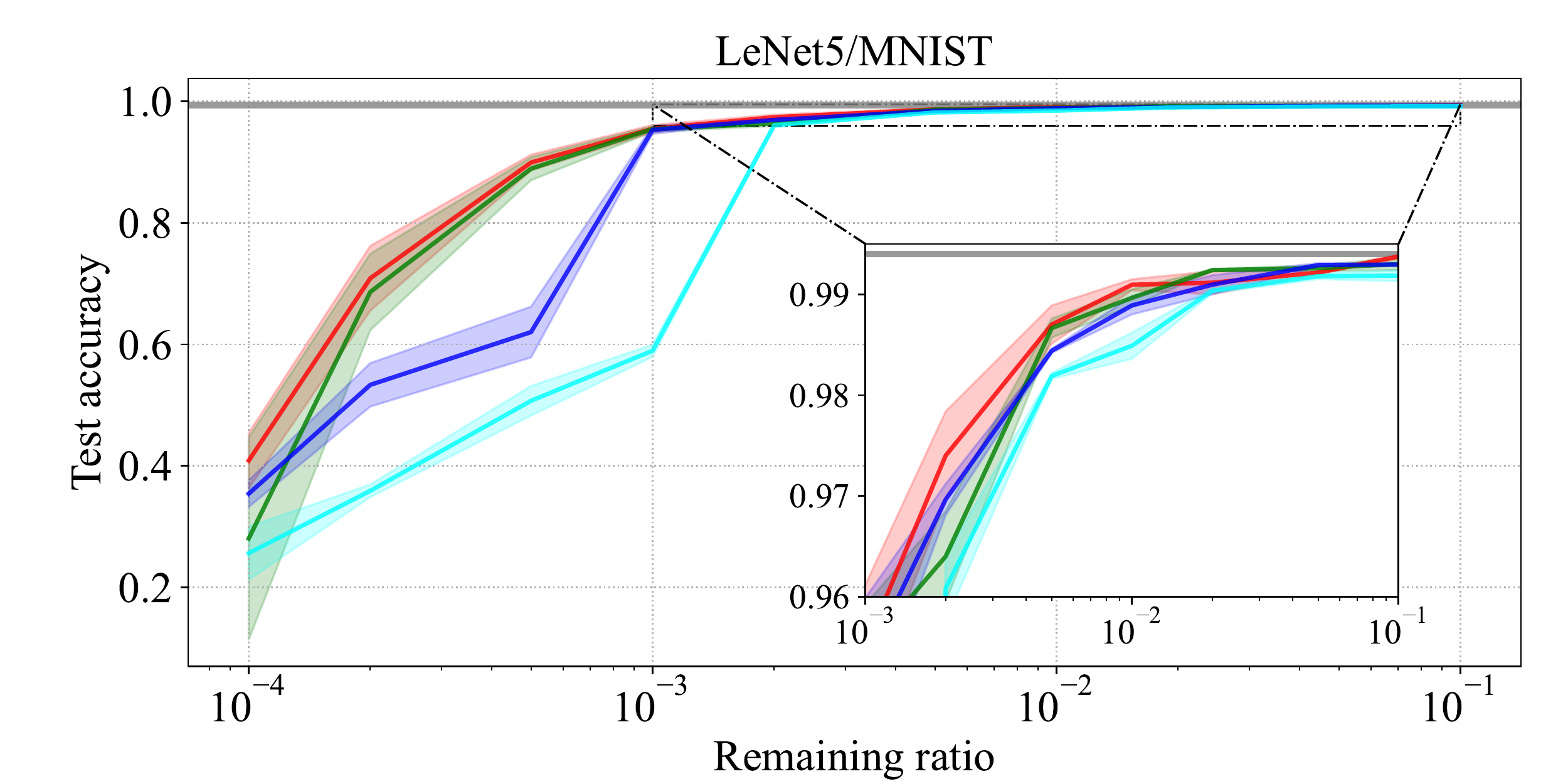}
    \end{subfigure}
    \begin{subfigure}{\textwidth}
    \includegraphics[width=\columnwidth]{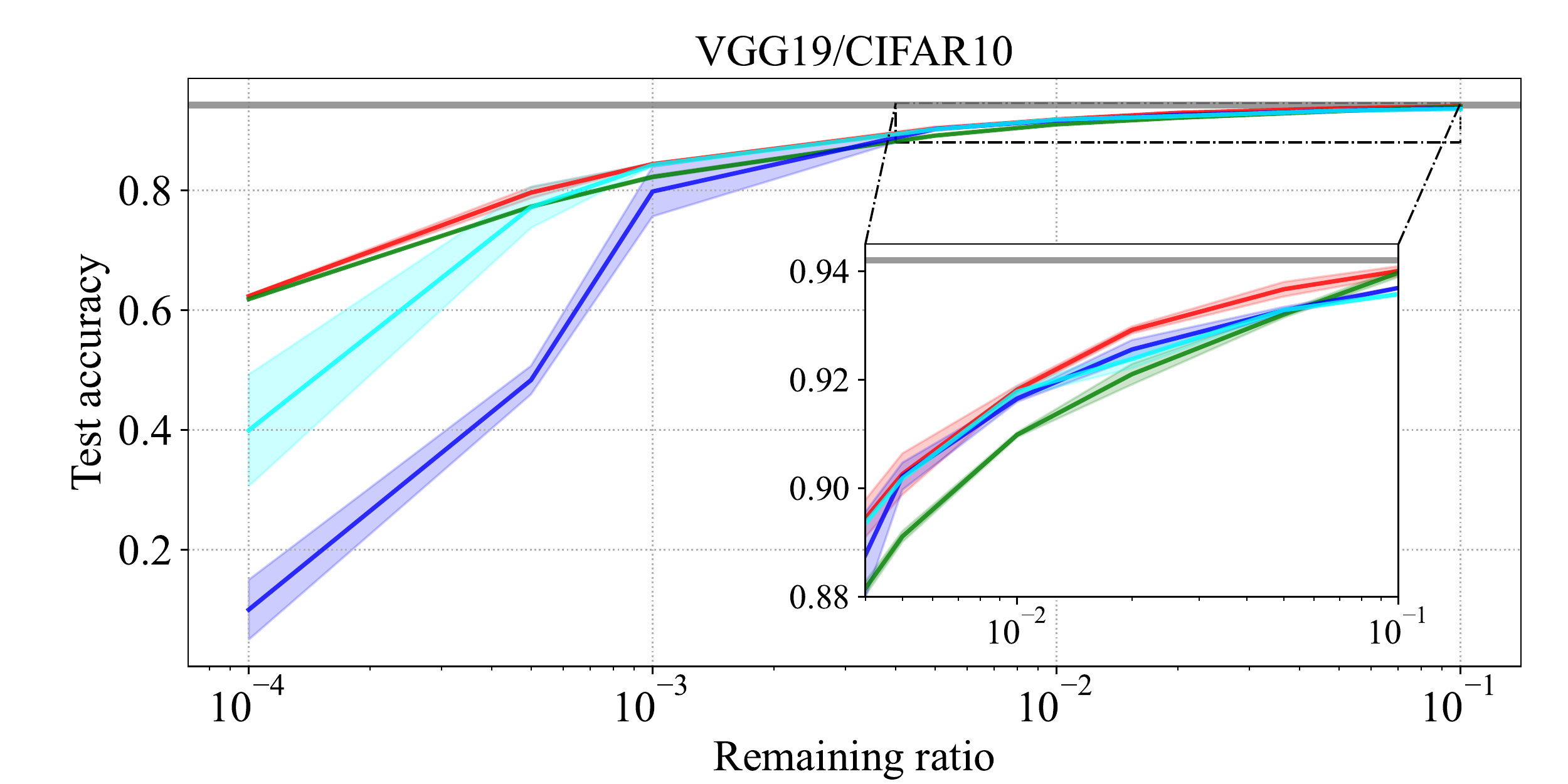}
    \end{subfigure}
    \begin{subfigure}{\textwidth}
    \includegraphics[width=\columnwidth]{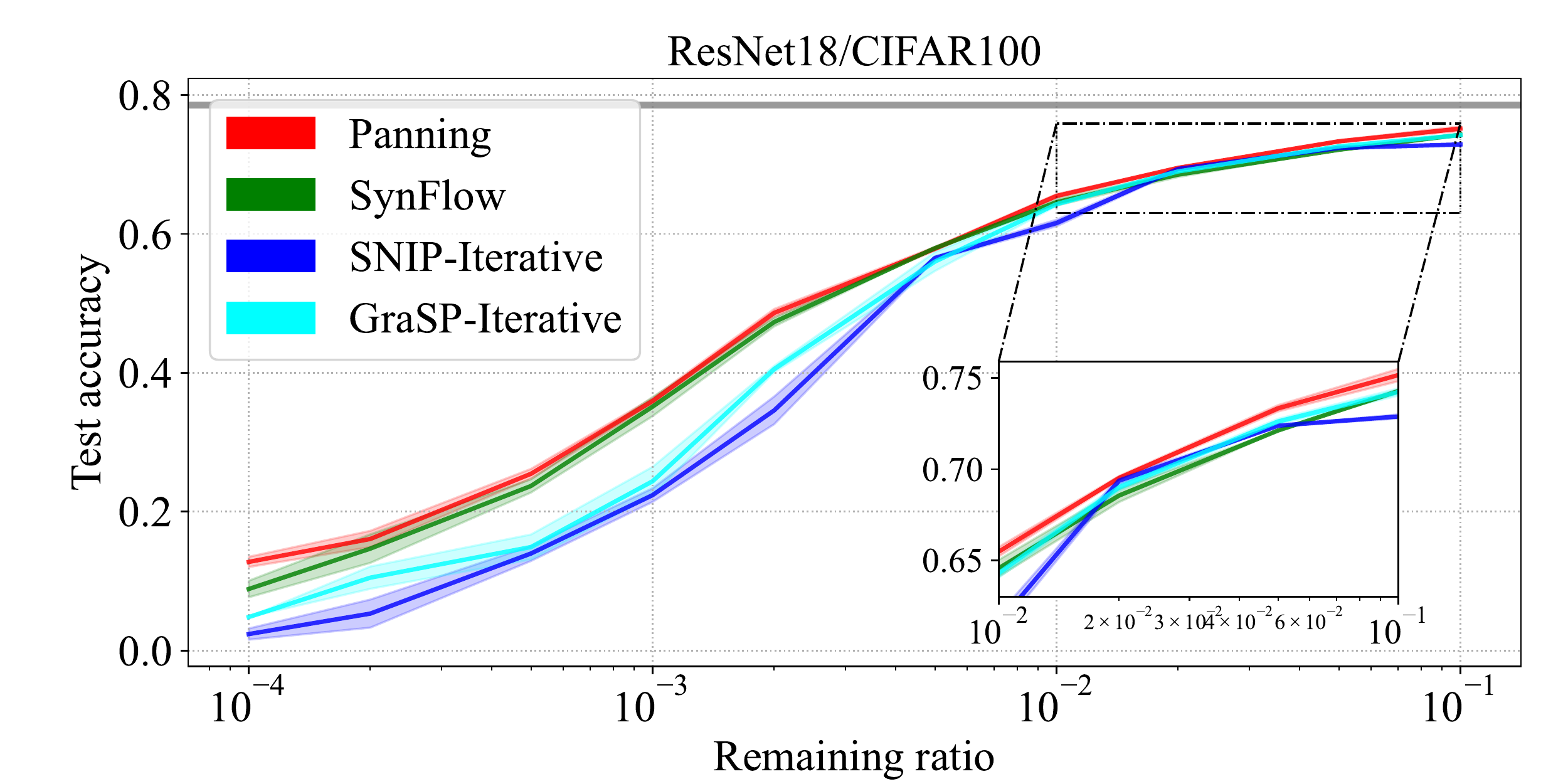}
    \end{subfigure}
    \caption{Performance comparison of Panning and iterative pruning. The shaded area indicates the error of repeated experiments. The gray horizontal line is the baseline.}
    \label{panning_performance}
  \endminipage\hfill
  \minipage{0.49\textwidth}
    \begin{subfigure}{\textwidth}
    \includegraphics[width=\columnwidth]{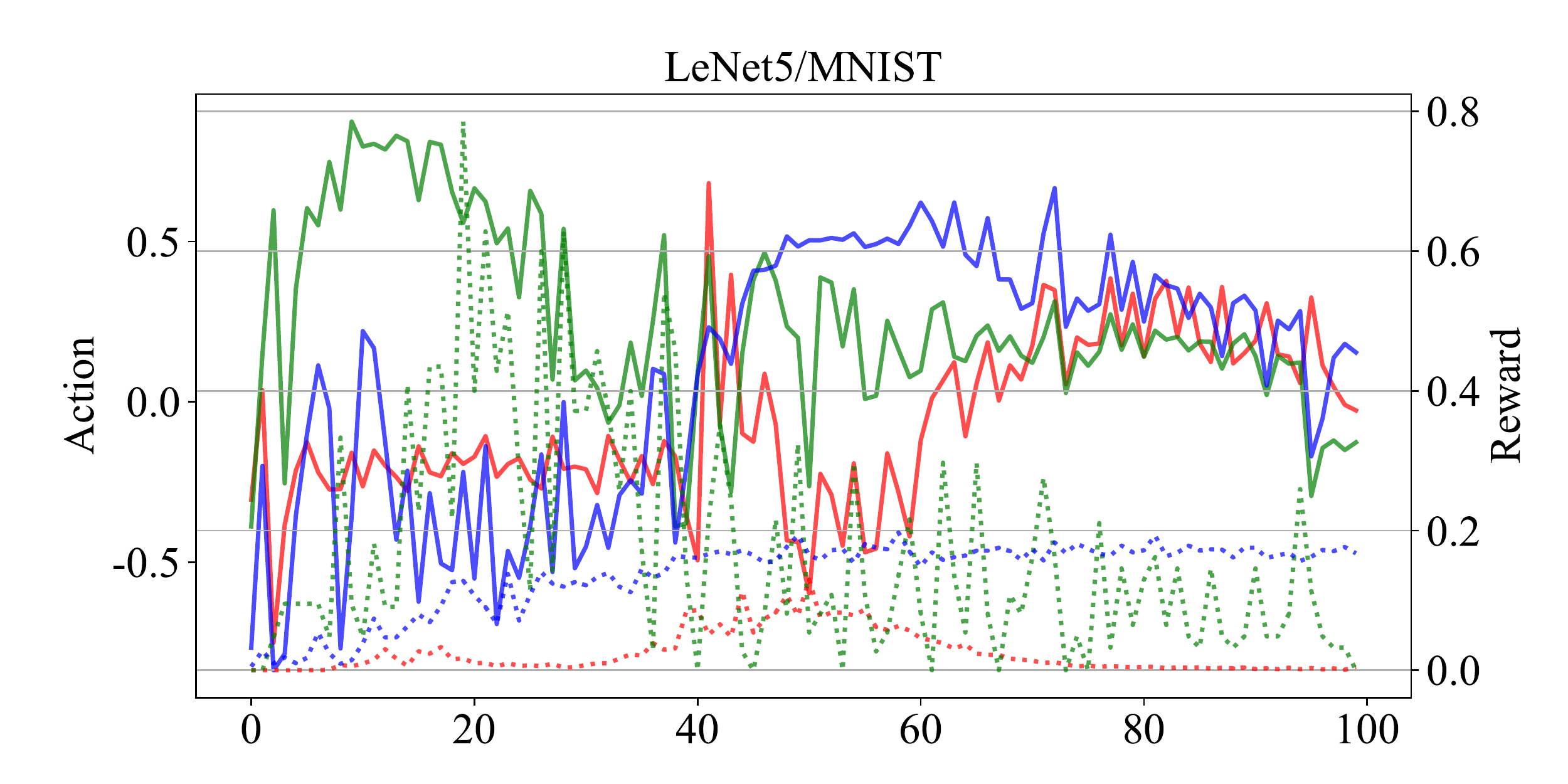}
    \end{subfigure}
    \begin{subfigure}{\textwidth}
    \includegraphics[width=\columnwidth]{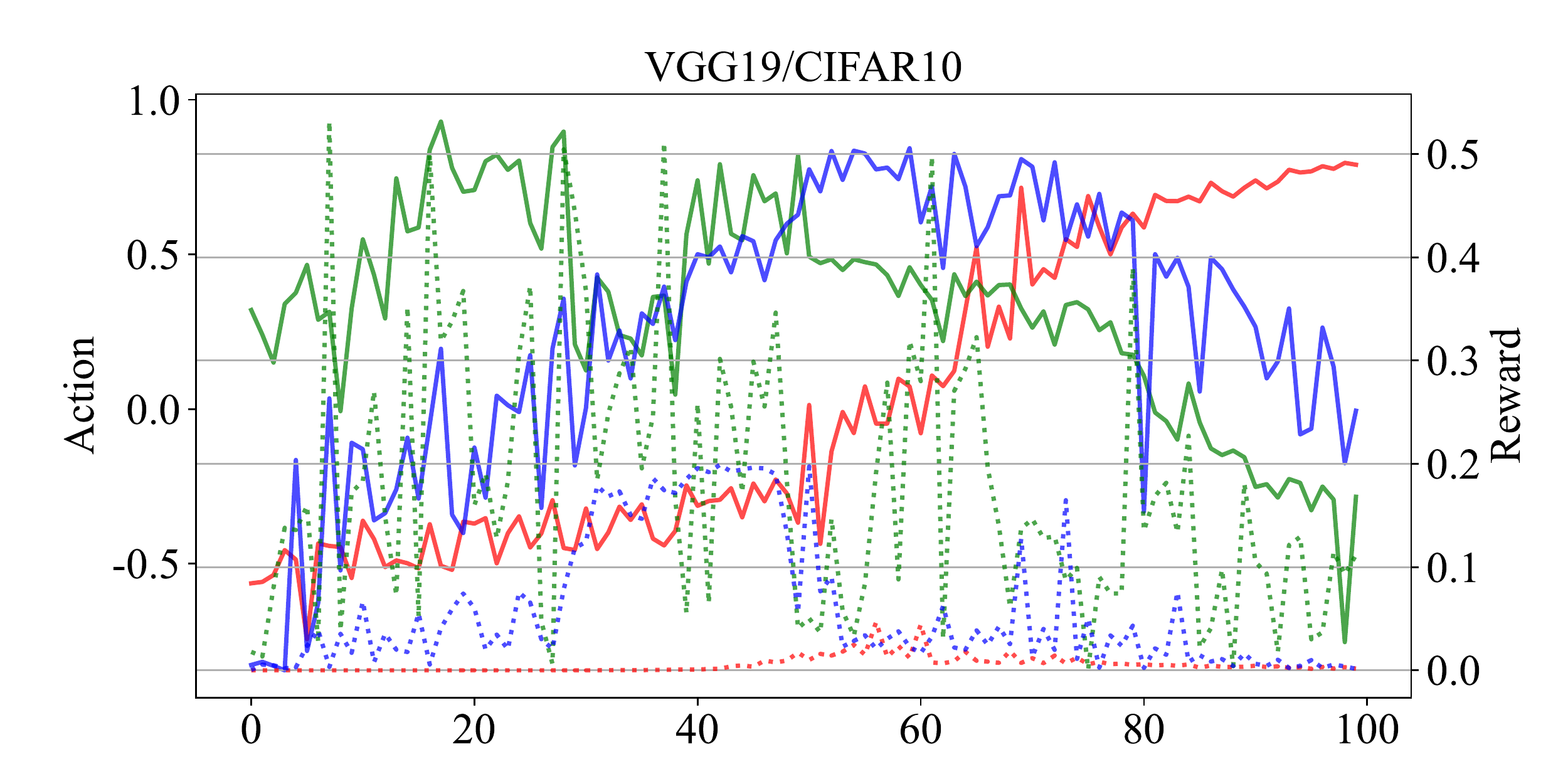}
    \end{subfigure}
    \begin{subfigure}{\textwidth}
    \includegraphics[width=\columnwidth]{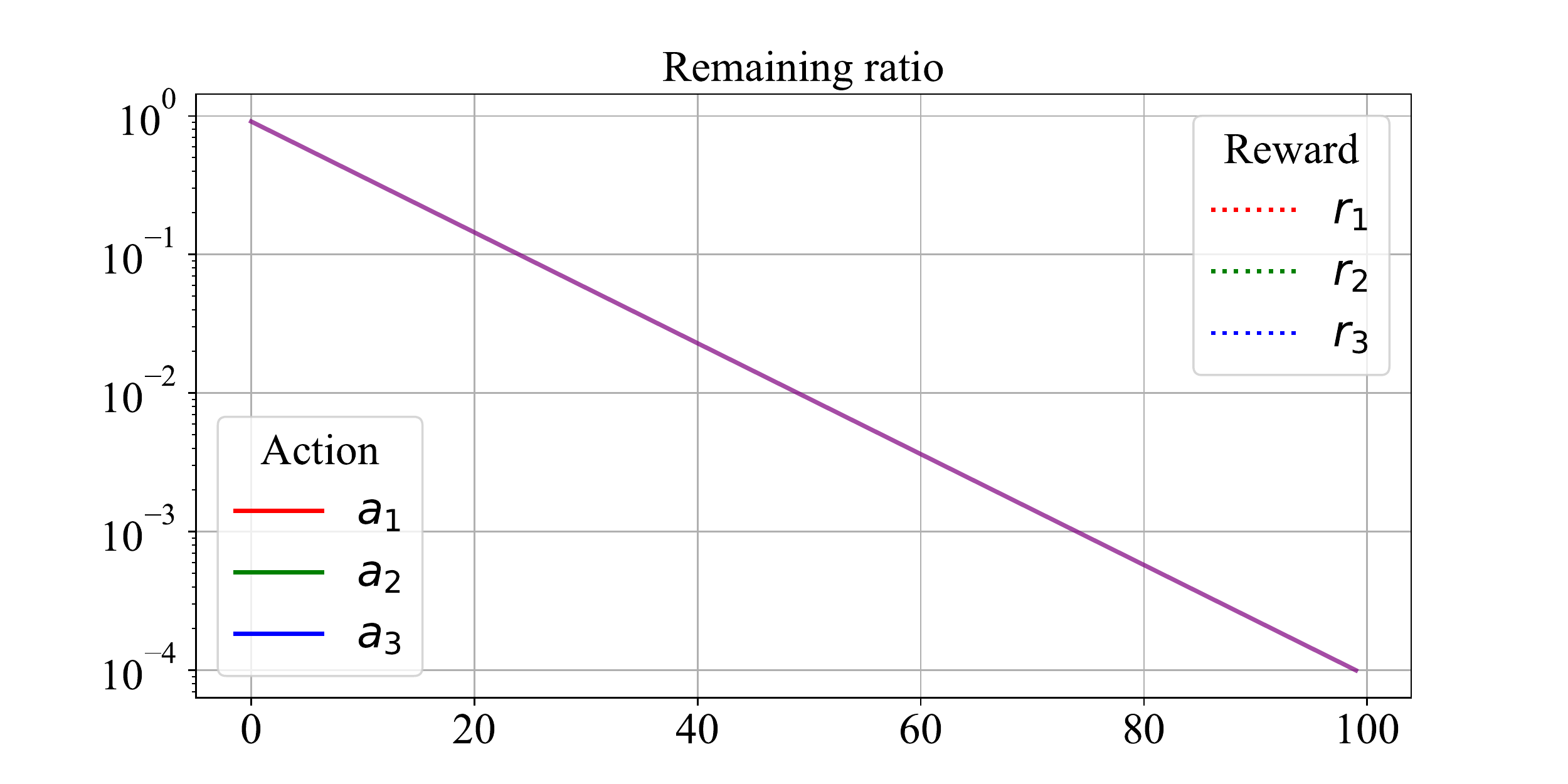}
    \end{subfigure}
    \caption{Action and reward changes generated by RLPanning during pruning. The left axis is the output action $a$ of the agent. The right axis is the reward $r$ of environmental feedback. The bottom graph shows the change in the remaining ratio over iterations.}
    \label{rlpanning_action}
  \endminipage
\end{figure*}

%-----------------------------------------------------------
\subsection{Experimental results of artificial Panning}

In \cref{metrics}, we introduce the metrics selected by Panning. In the process of Panning, when the compression rate is low, the balance between layers is better, and the weight sensitive to loss should have the highest priority. When the compression rate is high, the number of weights is minimal, and it is easy to have connectivity matters\cite{Connectivity21}, so the sensitivity between layers is more important. Combined with the influence of gradient norm on trainability, the hyperparameter settings under different compression stages in the Panning process are shown in \cref{panning_hyp}.

We compare the compression ratio of Panning, SynFlow, SNIP iteration, and GraSP iteration between \{$10^{-1},10^{-4}$\}. In order to highlight the effectiveness of Panning, the SNIP and GraSP iterations also use the FORCE\cite{Jorge21Trimming} method to prune the network dynamically, and the number $T$ of iterative pruning is unified as 100 times. Taking the L2 regularized complete network as the baseline, \cref{panning_performance} shows the experimental results on VGG19/CIFAR10, ResNet18/CIFAR100, and LeNet5/MNIST.

Note that the curve corresponding to Panning in the figure is always above the other methods, whether with a high or low compression ratio. The MNIST dataset is relatively simple, and the differences between the various methods are minor. The advantage of Panning on VGG19/CIFAR10 is mainly manifested in the low compression rate. On ResNet18/CIFAR100, the generalization ability through Panning is also maintained well at extremely high compression rates. In addition, the hyperparameters that adjust the proportion of metrics in Panning are set based on our estimates. In fact, if the parameters in \cref{panning_hyp} are further tuned, Panning will achieve better performance, but our estimated settings have demonstrated the effectiveness of Panning for multiple screening.

\begin{table}[t]
  \centering
  \caption{Hyperparameter setting of artificial Panning.}
  \label{panning_hyp}
  \resizebox{0.9\columnwidth}{!}{
  \renewcommand\arraystretch{0.8}
  \begin{tabular}{l | c c c}
  \cmidrule[\heavyrulewidth](lr){1-4}
  Pruning ratio & \makebox[0.17\textwidth]{$p_1$(SynFlow)} & \makebox[0.17\textwidth]{$p_2$(SNIP)} & \makebox[0.17\textwidth]{$p_3$(GraSP)} \\ 
  \cmidrule(lr){1-4}
  $(0,0.8]$ & 0.2 & 0.5 & 0.3 \\ 
  $(0.8,0.9]$ & 0.2 & 0.4 & 0.4 \\ 
  $(0.9,0.98]$ & 0.2 & 0.3 & 0.5 \\ 
  $(0.98,0.99]$ & 0.4 & 0.2 & 0.4 \\ 
  $(0.99,1)$ & 0.5 & 0 & 0.5 \\ 
  \cmidrule(lr){1-4}
  \end{tabular}}
\end{table}

\begin{table}[t]
  \centering
  \caption{TD3 training details and hyperparameters.}
  \label{rlpanning_hyp}
  \resizebox{0.95\columnwidth}{!}{
  \renewcommand\arraystretch{0.8}
  \begin{tabular}{lc | lc}
  \cmidrule[\heavyrulewidth](lr){1-4}
  % \makebox[0.17\textwidth]{Training} & \makebox[0.17\textwidth]{TD3} \\ 
  \multicolumn{2}{c|}{Training} & \multicolumn{2}{c}{TD3} \\ 
  \cmidrule(lr){1-4}
  \makebox[0.3\textwidth][l]{Optimizer} & Adam & \makebox[0.3\textwidth][l]{Exploration Noise} & 0.1 \\ 
  Learning Rate & 3e-4 & Discount Factor & 0.99 \\ 
  Start Timesteps & 2e3 & Network Update Rate & 0.01 \\ 
  Max Timesteps & 2e5 & Policy Noise & 0.2 \\ 
  Batch Size & 256 &   &   \\ 
  \cmidrule(lr){1-4}
  \end{tabular}}
\end{table}

\begin{table}[h]\scriptsize
  \centering
  \caption{Performance comparison of RLPanning and Panning.}
  \label{rlpanning_panning}
  \resizebox{0.9\columnwidth}{!}{
  \begin{tabular}{l cccccc}
  \cmidrule[\heavyrulewidth](lr){1-7}
  \makebox[0.3\columnwidth][l]{\textbf{LeNet5/MNIST}} & \multicolumn{6}{c}{Acc: 99.40\%} \\ 
  \cmidrule(lr){1-7}
  {Pruning ratio} & 90\% & 95\% & 98\% & 99\% & 99.9\% & 99.99\% \\ 
  \cmidrule(lr){1-7}
  Panning & 99.37 & \textbf{99.21} & 99.11 & 99.09 & 95.51 & 40.85 \\
  RLPanning	& \textbf{99.40} & 99.18 & \textbf{99.23} & \textbf{99.15} & \textbf{97.89} & \textbf{45.43} \\
  \end{tabular}}

  \resizebox{0.9\columnwidth}{!}{
  \begin{tabular}{l cccccc}
  \cmidrule[\heavyrulewidth](lr){1-7}
  \makebox[0.3\columnwidth][l]{\textbf{LeNet5/FashionMNIST}} & \multicolumn{6}{c}{Acc: 91.98\%} \\ 
  \cmidrule(lr){1-7}
  % {Pruning ratio} & 90\% & 95\% & 98\% & 99\% & 99.9\% & 99.99\% \\ 
  % \cmidrule[\heavyrulewidth](lr){1-7}
  Panning & 90.67 & 90.14 & 88.74 & 86.92 & 68.97 & 26.67 \\
  RLPanning	& \textbf{90.84} & \textbf{90.53} & \textbf{89.35} & \textbf{87.41} & \textbf{69.02} & \textbf{30.26} \\
  \end{tabular}}
  
  \resizebox{0.9\columnwidth}{!}{
  \begin{tabular}{l cccccc}
  \cmidrule[\heavyrulewidth](lr){1-7}
  \makebox[0.3\columnwidth][l]{\textbf{VGG19/CIFAR10}} & \multicolumn{6}{c}{Acc: 94.20\%} \\ 
  \cmidrule(lr){1-7}
  % {Pruning ratio} & 90\% & 95\% & 98\% & 99\% & 99.9\% & 99.99\% \\ 
  % \cmidrule[\heavyrulewidth](lr){1-7}
  Panning & 94.02 & 93.66 & 92.81 & 91.75 & 83.87 & 62.32 \\
  RLPanning	& \textbf{94.09} & \textbf{93.80} & \textbf{92.98} & \textbf{92.12} & \textbf{84.52} & \textbf{64.51} \\
  \cmidrule[\heavyrulewidth](lr){1-7}
  \end{tabular}}
\end{table}

\begin{table}[!]
  \centering
  \caption{Performance comparison of pruned VGG19 and ResNet18 on Tiny-ImageNet.}
  \label{vgg_resnet_tiny_imagenet}
  \resizebox{0.96\columnwidth}{!}{
  \renewcommand\arraystretch{0.9}
  \begin{tabular}{l ccc ccc}
  \cmidrule[\heavyrulewidth](lr){1-7}
  \textbf{Network} & 
  \multicolumn{3}{c}{\makebox[0.33\columnwidth]{VGG19: 63.29\%}} & 
  \multicolumn{3}{c}{\makebox[0.33\columnwidth]{ResNet18: 63.92\%}} \\ 
  \cmidrule(lr){1-1} \cmidrule(lr){2-4} \cmidrule(lr){5-7}
  Pruning ratio & 
  \makebox[0.11\columnwidth]{90\%} & \makebox[0.11\columnwidth]{95\%} & \makebox[0.11\columnwidth]{98\%} & 
  \makebox[0.11\columnwidth]{90\%} & \makebox[0.11\columnwidth]{95\%} & \makebox[0.11\columnwidth]{98\%} \\ 
  \cmidrule(lr){1-1} \cmidrule(lr){2-4} \cmidrule(lr){5-7}
  SNIP &	
  61.15 &	59.32 &	49.04 &	60.42 &	58.56 & 50.66 \\
  GraSP &	
  60.26 &	59.53 &	56.54 & 60.18 &	58.84 &	55.79 \\
  SynFlow &	
  59.54 &	58.06 &	45.29 & 59.03 &	56.77 &	46.34 \\
  Panning &	
  61.67 &	\textbf{60.25} &	\textbf{58.33} & 60.74 &	\textbf{58.92} &	\textbf{56.48} \\
  RLPanning	&
  \textbf{61.84} &	59.83 &	58.17 & \textbf{61.07} &	58.73 &	55.82 \\
  \midrule
  \end{tabular}}
\end{table}

%-----------------------------------------------------------
\subsection{Experimental results of RLPanning}

The actor and critic networks of the TD3 agent in RLPanning are both three-layer linear networks, using the Tanh activation function, and the dimension of the hidden layer is 256. The training details and key hyperparameters are shown in \cref{rlpanning_hyp}. When training the agent, we only use LeNet5/MNIST as the network and dataset for the Panning environment and do not need other network structures and datasets. The resulting Panning agent can be adapted to other convolutional network pruning tasks.

We apply the trained Panning agent to prune LeNet5 and VGG19. The feedback of the Panning environment and the actions of the Panning agent are shown in \cref{rlpanning_action}. Where $\{r_1,r_2,r_3\}$ correspond to the first three unsigned items in \cref{reward_eq}, respectively. The reward $r$ comes from the state of the sparse network, and the $r_2$ change is more drastic than other reward items due to different data batches. It is not difficult to find from the figure that the action $a_2$ plays a leading role in the early stage of the iteration. As the remaining weights decrease, the magnitudes of actions $a_1, a_3$ begin to increase. By the late iteration, the reward $r$ gradually stabilizes, with comparable action magnitude in LeNet5/MNIST and even much higher action $a_1$ in VGG19/CIFAR10, due to the deeper network layers of VGG and the more significant challenge of inter-layer equalization. Note that the actions produced by the agent are roughly consistent with our preset changes, and the same pattern occurs on untrained networks and datasets.

The performance of RLPanning pruned LeNet5 and VGG19 is shown in \cref{rlpanning_panning}. We take the average of three experiments to get the results. It is observed that there is less room for improvement at low compression ratios, and RLPanning is very close to Panning in terms of test accuracy. However, RLPanning still shows the advantage of dynamic pruning according to the environment at a high compression rate. \cref{vgg_resnet_tiny_imagenet} shows the performance comparison on TinyImageNet. It is observed that Panning has very good performance on large datasets, or rather more obvious improvement in the underfitting state. Note that RLPanning is analogous to Panning on TinyImageNet and significantly outperforms other methods. RLPanning is a little more prominent at low compression ratios. Note that the TD3 model we trained based on LeNet5/MNIST, and it is impossible to accurately capture features for large datasets. But even so, the experimental data verifies the effectiveness of the agent trained on the small model can still be applied to the large model and large dataset. This means that agents trained with reinforcement learning can suppress the interference caused by changes in dataset size and network scale, and can more profoundly reflect the inherent nature of weight screening. 

%===========================================================
\section{Conclusion}

In the experiments of this paper, we generalize the pruning process as the process of expressive force transfer, and analyze and improve the existing weight metrics before training. We then propose a pruning before training approach for Panning and automate the Panning process through reinforcement learning. Our experimental results show that Panning further improves the performance of pruned neural networks before training. The expressive force transfer is a fascinating phenomenon, and in the follow-up work, we will consider the higher-order terms of the Taylor approximation to design a new metric. In addition, we will make further explorations in structured sparsity and constraint-guided sparsity based on expressive force transfer.

\subsubsection{Acknowledgements} This research was partially supported by Hunan Provincial Key Laboratory of Intelligent Processing of Big Data on Transp. Thanks to Google Cloud and Huawei Cloud for cloud computing services.

% \bibliographystyle{unsrt}

\begin{thebibliography}{10}
\providecommand{\url}[1]{\texttt{#1}}
\providecommand{\urlprefix}{URL }
\providecommand{\doi}[1]{https://doi.org/#1}

\bibitem{DAM21}
Bu, J., Daw, A., Maruf, M., Karpatne, A.: Learning compact representations of
    neural networks using discriminative masking {(DAM)}. In: NeurIPS. pp.
    3491--3503 (2021)

\bibitem{ESPN}
Cho, M., Joshi, A., Hegde, C.: {ESPN:} extremely sparse pruned networks. CoRR
    \textbf{abs/2006.15741} (2020), \url{https://arxiv.org/abs/2006.15741}

\bibitem{desai2019evaluating}
Desai, S., Zhan, H., Aly, A.: Evaluating lottery tickets under distributional
    shifts. EMNLP-IJCNLP 2019 p.~153 (2019). \doi{10.18653/v1/D19-6117}

\bibitem{Lottery}
Frankle, J., Carbin, M.: The lottery ticket hypothesis: Finding sparse,
    trainable neural networks. In: {ICLR} (2019),
    \url{https://openreview.net/forum?id=rJl-b3RcF7}

\bibitem{Missing}
Frankle, J., Dziugaite, G.K., Roy, D., Carbin, M.: Pruning neural networks at
    initialization: Why are we missing the mark? In: {ICLR} (2021),
    \url{https://openreview.net/forum?id=Ig-VyQc-MLK}

\bibitem{frankle2019stabilizing}
Frankle, J., Dziugaite, G.K., Roy, D.M., Carbin, M.: Stabilizing the lottery
    ticket hypothesis. arXiv preprint arXiv:1903.01611  (2019),
    \url{https://doi.org/10.48550/arXiv.1903.01611}

\bibitem{TD3}
Fujimoto, S., van Hoof, H., Meger, D.: Addressing function approximation error
    in actor-critic methods. In: {ICML}. Proceedings of Machine Learning
    Research, vol.~80, pp. 1582--1591. {PMLR} (2018)

\bibitem{VACL19}
Gao, S., Liu, X., Chien, L., Zhang, W., Alvarez, J.M.: {VACL:} variance-aware
    cross-layer regularization for pruning deep residual networks. In: {ICCV}
    Workshops. pp. 2980--2988. {IEEE} (2019),
    \url{https://doi.org/10.1109/ICCVW.2019.00360}

\bibitem{goodfellow2016deep}
Goodfellow, I., Bengio, Y., Courville, A.: Deep learning. MIT press (2016)

\bibitem{Compression2016}
Han, S., Mao, H., Dally, W.J.: Deep compression: Compressing deep neural
    network with pruning, trained quantization and huffman coding. In: {ICLR}
    (2016)

\bibitem{Prune2015}
Han, S., Pool, J., Tran, J., Dally, W.J.: Learning both weights and connections
    for efficient neural network. In: {NIPS}. pp. 1135--1143 (2015)

\bibitem{Robust2021}
Hayou, S., Ton, J., Doucet, A., Teh, Y.W.: Robust pruning at initialization.
    In: {ICLR} (2021), \url{https://openreview.net/forum?id=vXj\_ucZQ4hA}

\bibitem{AMC18}
He, Y., Lin, J., Liu, Z., Wang, H., Li, L., Han, S.: {AMC:} automl for model
    compression and acceleration on mobile devices. In: {ECCV} {(7)}. Lecture
    Notes in Computer Science, vol. 11211, pp. 815--832. Springer (2018)

\bibitem{Jorge21Trimming}
de~Jorge, P., Sanyal, A., Behl, H.S., Torr, P.H.S., Rogez, G., Dokania, P.K.:
    Progressive skeletonization: Trimming more fat from a network at
    initialization. In: {ICLR} (2021),
    \url{https://openreview.net/forum?id=9GsFOUyUPi}

\bibitem{Alexnet2012}
Krizhevsky, A., Sutskever, I., Hinton, G.E.: Imagenet classification with deep
    convolutional neural networks. In: {NIPS}. pp. 1106--1114 (2012)

\bibitem{BERT2020}
Lan, Z., Chen, M., Goodman, S., Gimpel, K., Sharma, P., Soricut, R.: {ALBERT:}
    {A} lite {BERT} for self-supervised learning of language representations. In:
    {ICLR}. OpenReview.net (2020)

\bibitem{Lee20Signal}
Lee, N., Ajanthan, T., Gould, S., Torr, P.H.S.: A signal propagation
    perspective for pruning neural networks at initialization. In: {ICLR} (2020),
    \url{https://openreview.net/forum?id=HJeTo2VFwH}

\bibitem{SNIP}
Lee, N., Ajanthan, T., Torr, P.H.S.: Snip: single-shot network pruning based on
    connection sensitivity. In: {ICLR} (Poster) (2019),
    \url{https://openreview.net/forum?id=B1VZqjAcYX}

\bibitem{PruningFilters17}
Li, H., Kadav, A., Durdanovic, I., Samet, H., Graf, H.P.: Pruning filters for
    efficient convnets. In: {ICLR} (Poster). OpenReview.net (2017)

\bibitem{DDPG}
Lillicrap, T.P., Hunt, J.J., Pritzel, A., Heess, N., Erez, T., Tassa, Y.,
    Silver, D., Wierstra, D.: Continuous control with deep reinforcement
    learning. In: {ICLR} (Poster) (2016)

\bibitem{Towards19}
Lin, S., Ji, R., Yan, C., Zhang, B., Cao, L., Ye, Q., Huang, F., Doermann,
    D.S.: Towards optimal structured {CNN} pruning via generative adversarial
    learning. In: {CVPR}. pp. 2790--2799. Computer Vision Foundation / {IEEE}
    (2019)

\bibitem{OverParame21}
Liu, S., Yin, L., Mocanu, D.C., Pechenizkiy, M.: Do we actually need dense
    over-parameterization? in-time over-parameterization in sparse training. In:
    {ICML}. Proceedings of Machine Learning Research, vol.~139, pp. 6989--7000.
    {PMLR} (2021)

\bibitem{LearningEfficient17}
Liu, Z., Li, J., Shen, Z., Huang, G., Yan, S., Zhang, C.: Learning efficient
    convolutional networks through network slimming. In: {ICCV}. pp. 2755--2763.
    {IEEE} Computer Society (2017)

\bibitem{Rethinking19}
Liu, Z., Sun, M., Zhou, T., Huang, G., Darrell, T.: Rethinking the value of
    network pruning. In: {ICLR} (Poster) (2019),
    \url{https://openreview.net/forum?id=rJlnB3C5Ym}

\bibitem{Malach2020Proving}
Malach, E., Yehudai, G., Shalev{-}Shwartz, S., Shamir, O.: Proving the lottery
    ticket hypothesis: Pruning is all you need. In: {ICML}. Proceedings of
    Machine Learning Research, vol.~119, pp. 6682--6691. {PMLR} (2020),
    \url{http://proceedings.mlr.press/v119/}

\bibitem{DQN}
Mnih, V., Kavukcuoglu, K., Silver, D., Rusu, A.A., Veness, J., Bellemare, M.G.,
    Graves, A., Riedmiller, M.A., Fidjeland, A., Ostrovski, G., Petersen, S.,
    Beattie, C., Sadik, A., Antonoglou, I., King, H., Kumaran, D., Wierstra, D.,
    Legg, S., Hassabis, D.: Human-level control through deep reinforcement
    learning. Nat.  \textbf{518}(7540),  529--533 (2015)

\bibitem{Scalable2018}
Mocanu, D.C., Mocanu, E., Stone, P., Nguyen, P.H., Gibescu, M.: Scalable
    training of artificial neural networks with adaptive sparse connectivity
    inspired by network science. Nature Communications  (2018).
    \doi{10.1038/s41467-018-04316-3}

\bibitem{morcos2019oneticket}
Morcos, A.S., Yu, H., Paganini, M., Tian, Y.: One ticket to win them all:
    generalizing lottery ticket initializations across datasets and optimizers.
    In: NeurIPS. pp. 4933--4943 (2019),
    \url{https://proceedings.neurips.cc/paper/2019/hash/a4613e8d72a61b3b69b32d040f89ad81-Abstract.html}

\bibitem{Logarithmic20}
Orseau, L., Hutter, M., Rivasplata, O.: Logarithmic pruning is all you need.
    In: NeurIPS (2020),
    \url{https://proceedings.neurips.cc/paper/2020/hash/1e9491470749d5b0e361ce4f0b24d037-Abstract.html}

\bibitem{CurriculumLearning18}
Qu, M., Tang, J., Han, J.: Curriculum learning for heterogeneous star network
    embedding via deep reinforcement learning. In: {WSDM}. pp. 468--476. {ACM}
    (2018)

\bibitem{LargeScale17}
Real, E., Moore, S., Selle, A., Saxena, S., Suematsu, Y.L., Tan, J., Le, Q.V.,
    Kurakin, A.: Large-scale evolution of image classifiers. In: {ICML}.
    Proceedings of Machine Learning Research, vol.~70, pp. 2902--2911. {PMLR}
    (2017)

\bibitem{RandomTickets}
Su, J., Chen, Y., Cai, T., Wu, T., Gao, R., Wang, L., Lee, J.D.:
    Sanity-checking pruning methods: Random tickets can win the jackpot. In:
    NeurIPS (2020)

\bibitem{SynFlow}
Tanaka, H., Kunin, D., Yamins, D.L., Ganguli, S.: Pruning neural networks
    without any data by iteratively conserving synaptic flow. In: NeurIPS (2020),
    \url{https://proceedings.neurips.cc/paper/2020/hash/46a4378f835dc8040c8057beb6a2da52-Abstract.html}

\bibitem{Connectivity21}
Vysogorets, A., Kempe, J.: Connectivity matters: Neural network pruning through
    the lens of effective sparsity (2021). \doi{10.48550/ARXIV.2107.02306}

\bibitem{GraSP}
Wang, C., Zhang, G., Grosse, R.B.: Picking winning tickets before training by
    preserving gradient flow. In: {ICLR} (2020),
    \url{https://openreview.net/forum?id=SkgsACVKPH}

\bibitem{LearningStructured16}
Wen, W., Wu, C., Wang, Y., Chen, Y., Li, H.: Learning structured sparsity in
    deep neural networks. In: {NIPS}. pp. 2074--2082 (2016)

\bibitem{ToPrune18}
Zhu, M., Gupta, S.: To prune, or not to prune: Exploring the efficacy of
    pruning for model compression. In: {ICLR} (Workshop). OpenReview.net (2018)

\bibitem{ArchitectureSearch17}
Zoph, B., Le, Q.V.: Neural architecture search with reinforcement learning. In:
    {ICLR}. OpenReview.net (2017)

\bibitem{Learning18}
Zoph, B., Vasudevan, V., Shlens, J., Le, Q.V.: Learning transferable
    architectures for scalable image recognition. In: {CVPR}. pp. 8697--8710.
    Computer Vision Foundation / {IEEE} Computer Society (2018)

\bibitem{ObjectDetection2019}
Zou, Z., Shi, Z., Guo, Y., Ye, J.: Object detection in 20 years: {A} survey.
    CoRR  \textbf{abs/1905.05055} (2019)

\end{thebibliography}

\end{document}